\documentclass[sigconf]{acmart}

\AtBeginDocument{%
  }

\setcopyright{acmlicensed}
\copyrightyear{2018}
\acmYear{2018}
\acmDOI{10.1145/3746027.3754595}
\acmISBN{979-8-4007-2035-2/2025/10}

\usepackage{hyperref}
\usepackage{url}
\usepackage{microtype}
\usepackage{graphicx}
\usepackage{amsmath}
\usepackage{mathtools}
\usepackage{amsthm}
\usepackage{balance}
\usepackage{pifont}   

\usepackage{tabu}
\usepackage{multirow}
\usepackage{tabularx}
\usepackage{array}

\usepackage[utf8]{inputenc}
\usepackage{booktabs}
\usepackage{pifont}

\theoremstyle{plain}
\newtheorem{theorem}{Theorem}[section]

\theoremstyle{definition}
\newtheorem{definition}[theorem]{Definition}

\theoremstyle{remark}

\usepackage{subcaption}
\usepackage{bm}

\usepackage{algorithm}
\usepackage{algorithmic}

\definecolor{Green}{rgb}{0,0.6,0.9}

\definecolor{Green}{rgb}{0,0.6,0.9}

\newcommand{\cmark}{\ding{51}}%
\newcommand{\xmark}{\ding{55}}%
 \definecolor{mygray}{gray}{0.6}

\usepackage{tcolorbox} 
\tcbuselibrary{theorems} 
\definecolor{mytheoremfr}{RGB}{200,200,200} 
\definecolor{mytheorembg}{RGB}{240,240,240} 

\newtcbtheorem{Prompt}{Prompt}{colback=mytheorembg!25,colframe=mytheoremfr, boxrule=0.5pt}{th}

\setcopyright{acmlicensed}
\acmDOI{10.1145/3746027.3754595}

\copyrightyear{2025}
\acmYear{2025}
\setcopyright{acmlicensed}\acmConference[MM'25]{Proceedings of the 33th ACM International Conference on Multimedia}{October 27--31, 2025}{Dublin, Ireland}
\acmBooktitle{Proceedings of the 33th ACM International Conference on Multimedia (MM '25), October 27--31, 2025, Dublin, Ireland}

\begin{document}

\title{PurifyGen: A Risk-Discrimination and Semantic-Purification Model for Safe Text-to-Image Generation}

\author{Zongsheng Cao}
\authornote{Equal contribution}
\email{agiczsr@gmail.com}
\affiliation{%
  \institution{Researcher}
  \country{}
}

\author{Yangfan He}
\authornotemark[1]
\email{he000577@umn.edu}
\affiliation{%
  \institution{UMN 
  }
  \country{}
  }

\author{Anran Liu}
\email{anniegogo1008@gmail.com}
\authornote{Corresponding author}
\affiliation{%
  \institution{Researcher}
  \country{}
}

  \author{Jun Xie} 
  \email{xiejun@lenovo.com}
  \affiliation{%
    \institution{PCIE}
    \country{}
  }
  
  \author{Feng Chen}
  \email{chenfeng@lenovo.com}
  \affiliation{
    \institution{PCIE}
    \country{}
  }

  \author{Zepeng Wang}  
  \authornotemark[2]
  \email{wangzpb@lenovo.com}
  \affiliation{%
    \institution{PCIE}
    \country{}
  }

\renewcommand{\shortauthors}{Zongsheng Cao et al.}

\begin{abstract}
  Recent advances in diffusion models have notably enhanced text-to-image (T2I) generation quality, but they also raise the risk of generating unsafe content. Traditional safety methods like text blacklisting or harmful content classification have significant drawbacks: they can be easily circumvented or require extensive datasets and extra training. To overcome these challenges, we introduce PurifyGen, a novel, training-free approach for safe T2I generation that retains the model's original weights. PurifyGen introduces a dual-stage strategy for prompt purification. First, we evaluate the safety of each token in a prompt by computing its complementary semantic distance, which measures the semantic proximity between the prompt tokens and concept embeddings from predeﬁned toxic and clean lists. This enables fine-grained prompt classification without explicit keyword matching or retraining. Tokens closer to toxic concepts are flagged as risky. Second, for risky prompts, we apply a dual-space transformation: we project toxic-aligned embeddings into the null space of the toxic concept matrix, effectively removing harmful semantic components, and simultaneously align them into the range space of clean concepts. This dual alignment purifies risky prompts by both subtracting unsafe semantics and reinforcing safe ones, while retaining the original intent and coherence. We further define a token-wise strategy to selectively replace only risky token embeddings, ensuring minimal disruption to safe content. PurifyGen offers a plug-and-play solution with theoretical grounding and strong generalization to unseen prompts and models. Extensive testing shows that PurifyGen surpasses current methods in reducing unsafe content across five datasets and competes well with training-dependent approaches. The code can refer to https://github.com/AI-Researcher-Team/PurifyGen.
\end{abstract}





\maketitle

\section{Introduction}

  In just a few years, generative AI has evolved from simple text completion into a comprehensive creative toolkit that now powers diverse media pipelines.  Breakthroughs first showcased by large-scale language models~\cite{brown2020language} have since migrated to automated program synthesis~\cite{chen2021evaluating}, fully synthetic audio tracks~\cite{kreuk2022audiogen,copet2024simple}, photorealistic image fabrication~\cite{podell2023sdxl,ho2022imagen}, and the generation of temporally coherent video sequences~\cite{ho2022imagen,kondratyuk2023videopoet,yoon2024raccoon,2024sora}.  Flagship systems such as \textit{DALL·E 3}~\cite{2023dalle3} and OpenAI’s \textit{Sora}~\cite{2024sora} have ignited fresh workflows across digital illustration, augmented/virtual reality, and educational media production.  At the same time, these models carry a darker potential: they can amplify societal bias, propagate discriminatory narratives, or produce explicit and violent visuals, prompting urgent debate on guardrails and responsible deployment.
  Breakthroughs first showcased by large-scale language models~\cite{brown2020language} have since migrated to automated program synthesis~\cite{chen2021evaluating}, fully synthetic audio tracks~\cite{kreuk2022audiogen,copet2024simple}, photorealistic image fabrication~\cite{podell2023sdxl,ho2022imagen}, and the generation of temporally coherent video sequences~\cite{ho2022imagen,kondratyuk2023videopoet,yoon2024raccoon,2024sora}.  Flagship systems such as \textit{DALL·E 3}~\cite{2023dalle3} and OpenAI's \textit{Sora}~\cite{2024sora} have ignited fresh workflows across digital illustration, augmented/
  virtual reality, and educational media production.  At the same time, these models carry a darker potential: they can amplify societal bias, propagate discriminatory narratives, or produce explicit and violent visuals, prompting urgent debate on guardrails and responsible deployment.

   \begin{figure}[ht]
    \small
    \centering
        \includegraphics[width=0.95\linewidth]{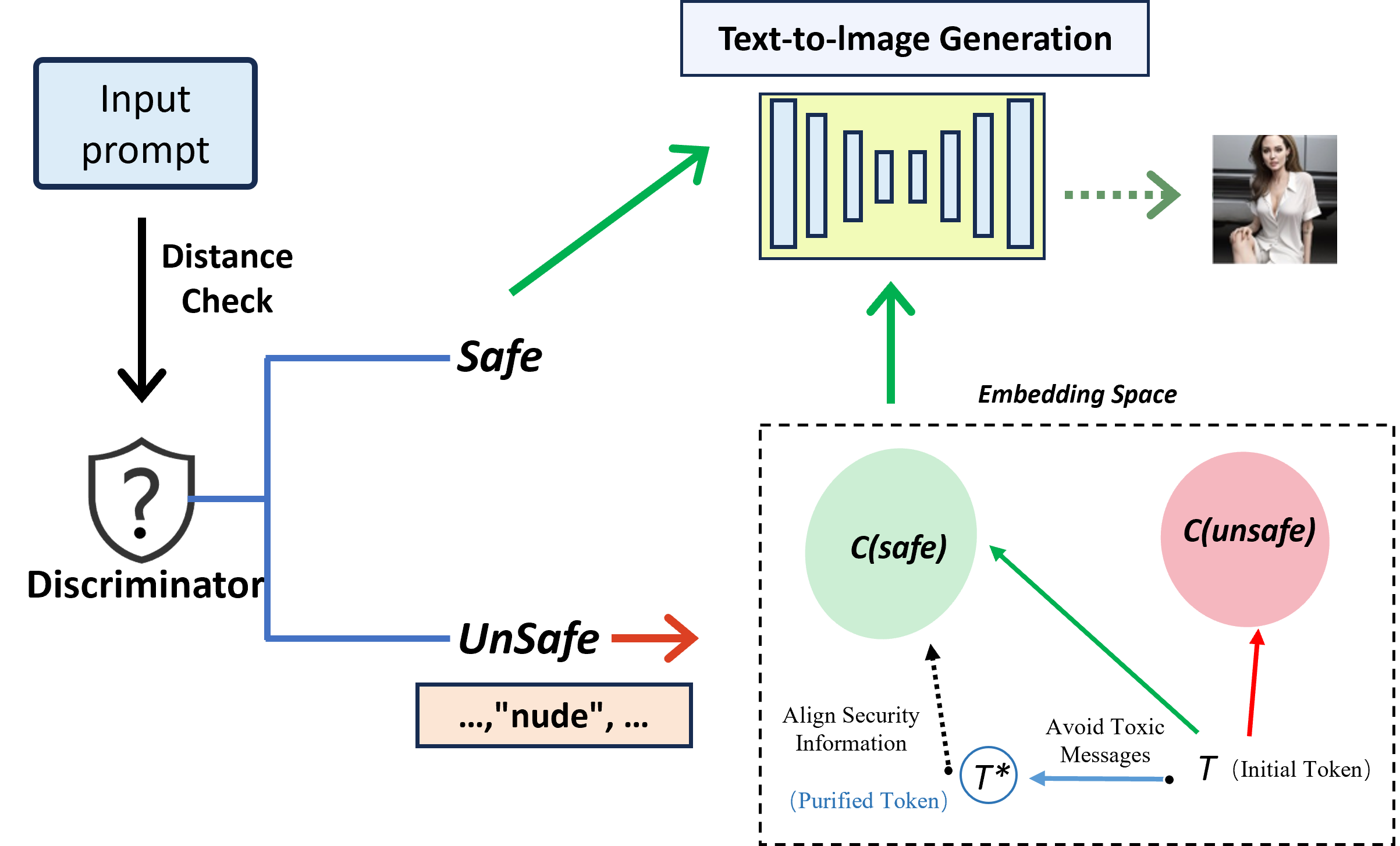}
        \caption{\small We present PurifyGen, a risk-discrimination and semantic-purification model for T2I that \textbf{identifies and filters} out a variety of user-defined concepts. Unlike the previous approach of filtering prompts directly, which may impair the semantics of normal prompts, PurifyGen enables the safe and faithful generation in an interpretable manner that can remove toxic concepts and create a safer version of inappropriate prompts without requiring any model updates.}
        \label{fig:teaser}
    \end{figure}
  
   Recent efforts to render text-to-image (T2I) generators safer have coalesced around two divergent toolkits.  
   Unlearning methods deliberately push the network through another round of training so that it “forgets’’ objectionable visual patterns altogether~\cite{zhang2023forget,huang2023receler,park2024direct,wu2024unlearning}.  While this can scrub out unwanted content, the procedure is computationally heavy and must be repeated whenever a new harmful motif surfaces.  
   Running on a different track, model-editing algorithms perform pinpoint parameter surgery to steer generation away from unsafe regions of the data manifold~\cite{orgad2023editing,gandikota2024unified,xiong2024editing}. These edits often carry a cost: image ﬁdelity may degrade, or the model’s broader capabilities may become destabilized, making it difﬁcult to remove toxic elements while preserving the user’s original intent.

  An emerging, more promising line of research focuses on training-free, filtering-based approaches. These methods aim to filter and modify input prompts to prevent unsafe content generation while preserving the model's original generative capabilities. Despite their appeal, existing filtering techniques face two critical limitations. First, they lack a nuanced risk-discrimination mechanism: by employing a hard filtering strategy that uniformly alters the input prompt, they fail to identify and target only the risky components, often leading to inefficient processing and unnecessary semantic loss. Second, these methods do not incorporate a mechanism for semantic purification, making it difficult to precisely cleanse harmful elements without disrupting the intended prompt meaning.
  In light of these deficiencies, it is natural to ask: \textit{Can we develop a unified framework that both discriminates between safe and risky prompt components and effectively sanitizes the unsafe portions while preserving the original intent?}

  To overcome the limitations of existing safety approaches in text-to-image generation, we propose \textbf{PurifyGen}, a training-free, adaptive, and plug-and-play framework that ensures safe generation without modifying the pretrained weights of diffusion models. Unlike retraining-based or rule-based systems, PurifyGen operates entirely in the prompt embedding space, making it highly efficient, interpretable, and broadly compatible across model architectures.
  Building such a framework requires addressing two core challenges:  
  \textbf{(C1)} Prompts often contain a mixture of safe and unsafe semantics, demanding fine-grained, token-level identification of risk rather than coarse, prompt-level decisions.  
  \textbf{(C2)} Removing risky semantics must be done without distorting the prompt's intended meaning, requiring a purification strategy that both filters out harmful components and restores alignment with safe content.
  
  To tackle \textbf{(C1)}, PurifyGen introduces a new semantic risk assessment mechanism. We compute the complementary semantic distance between prompt token embeddings and semantic concept spaces to quantify their relative association with safe or unsafe meanings. This enables fine-grained, token-wise risk discrimination that is generalizable and does not rely on hand-crafted keyword matching.
  To address \textbf{(C2)}, we propose a novel dual-space purification strategy. Risky tokens are projected into the null space of unsafe semantics to remove harmful content and simultaneously aligned into the range space of clean semantics to reinforce safe meaning. This transformation is performed selectively at the token level, ensuring that the remaining content remains coherent and faithful to the original prompt. Extensive experiments demonstrate that PurifyGen achieves state-of-the-art performance on five widely used safe T2I benchmarks: I2P~\cite{schramowski2023safe}, P4D~\cite{chin2024prompting4debugging}, Ring-A-Bell~\cite{tsai2024ring}, MMA-Diffusion~\cite{yang2024mma}, and UnlearnDiff~\cite{zhang2023generate}. It outperforms other training-free methods and delivers competitive results compared to top training-based methods.
  
  Our contributions are threefold:  
\begin{itemize}  
  \item We propose PurifyGen, a novel training-free framework for safe text-to-image generation that performs fine-grained, token-level risk assessment. By employing complementary semantic distance with user-defined blacklist and whitelist embeddings, PurifyGen enables interpretable and adaptive identification of unsafe prompt components without requiring model retraining or handcrafted keyword filtering.  

  \item We introduce a dual-space purification mechanism that combines null space projection to suppress toxic semantics and range space alignment to enhance safe content. This geometry-driven approach enables selective semantic editing at the embedding level, ensuring both safety and semantic coherence. The method is model-agnostic and can be integrated seamlessly as a plug-in module for various diffusion-based generative models.  

  \item We demonstrate that PurifyGen achieves state-of-the-art performance across public safety benchmarks, outperforming existing training-free methods and performing competitively against fine-tuned baselines. Our method exhibits strong generalization to unseen prompts, concepts, and model architectures, making it a practical and scalable solution for safety-critical generative applications.
\end{itemize}

  \section{Related Work}
  \subsection{T2I attacks}
  Recent studies have highlighted vulnerabilities in generative models, including large language models (LLMs)~\citep{patilcan, liu2024rethinking}, vision-language models (VLMs)~\citep{zhao2024evaluating}, and text-to-image (T2I) models~\citep{yang2024guardt2i, wang2024t2ishield, li2024safegen}. One notable vulnerability is cross-modality jailbreaks, as demonstrated by \cite{shayegani2023jailbreak}, which combine adversarial images with prompts to bypass VLM safeguards without directly targeting the language model. Tools such as Ring-A-Bell~\citep{tsai2024ring} and automated adversarial frameworks developed by \cite{kim2024automatic} and \cite{li2024art} focus on model-agnostic red-teaming techniques and adversarial prompt generation, uncovering safety weaknesses. Additionally, techniques by \cite{ma2024jailbreaking}, \cite{yang2024mma}, and \cite{mehrabi2023flirt} exploit text embeddings and multimodal inputs to evade model protections, employing strategies like adversarial prompt optimization and in-context learning~\citep{chin2024prompting4debugging, liu2024groot}. Collectively, these works emphasize the significant security vulnerabilities present in T2I models.

  \subsection{Safe T2I generation}
 Safe T2I has attract extensive attention in the past and there are mainly two branches for the safe T2I generation  as follow.

 Most prior attempts to curb unsafe or undesirable content intervene inside the generative model.  
 Early work concentrates on excising objectionable visual or textual patterns from the network through fine-tuning or negative guidance.  For example, \cite{zhang2024steerdiff} enlarges the safety margin by stripping away hazardous motifs, while \cite{li2024safegen} and \cite{gandikota2023erasing} demonstrate similar effects with concept erasure.  
 A complementary line of inquiry frames the problem adversarially: \cite{kim2024race} suppresses harmful embeddings via a race-conditioned discriminator, whereas \cite{das2024espresso,park2024direct} removes disallowed representations or optimizes preference scores to steer the generator away from toxic regions.  
 Beyond these, researchers have explored targeted parameter updates, such as cross-attention refinement \citep{lu2024mace}, continual-learning filters \citep{heng2023selective}, and latent-space surgery enabled by self-supervision \citep{liu2024latent,li2024self}.  
 Although these approaches can be effective, they usually require extensive retraining, incur notable computational overhead, and often introduce collateral degradation in visual quality.  
  
The second is the training-free paradigm. Safety mechanisms that sidestep any full re-training can be grouped into two broad camps. The first is closed-form weight editing techniques, where closed-form adjustments are computed once and grafted onto the generator's parameters to excise disallowed content while leaving its creative range intact.  Examples include model-projection editing \cite{gandikota2024unified}, target-embedding surgery \cite{gong2024reliable}, and the micro-update recipe of \cite{orgad2023editing} for diffusion backbones. A second line of work keeps the weights frozen and instead steers generation during inference through Safe Latent Diffusion with classifier-free guidance~\cite{schramowski2023safe}, or via prompt-level refinements in the Ethical Diffusion framework~\cite{cai2024ethical}.  
  Both families, however, have struggled to retain reliability once deployed on previously unseen prompts.  
  Our own approach adopts a risk-aware, semantic purification pipeline: it screens and sanitises the output on the fly, adapting its filters to the user's query without touching the underlying weights, thereby achieving stronger robustness and far better scalability in practice.

  \section{Preliminary}

  \subsection{Latent Diffusion Models}
  Diffusion generators~\cite{ho2020denoising} tackle image synthesis by beginning with pure Gaussian noise and then iteratively cleaning it across $T$ discrete steps until a coherent picture emerges.  At each timestep $t$, the network is tasked with predicting the residual noise still present, effectively acting as a time-conditioned denoiser.  Latent diffusion refines this recipe by performing the entire noise-to-signal procedure in a compact latent manifold and only decoding the final latent tensor back to pixel space at the end~\cite{rombach2022high}.  When extended to text-to-image generation, the model’s denoising function is additionally conditioned on a user prompt $p$. For instance, via classifier-free guidance~\cite{ho2022classifier}, the evolving sample gravitates toward an image whose semantics match the supplied description.

  Let $\epsilon_\theta(z_t, t, p)$ denote the noise estimation function parameterized by $\theta$. At time step $t$ ($0 < t \leq T$), the guided noise estimate is given by:
  \begin{equation}
  \hat{\epsilon}_\theta(z_t, t, p) = \epsilon_\theta(z_t, t, \varnothing) + \gamma \left( \epsilon_\theta(z_t, t, p) - \epsilon_\theta(z_t, t, \varnothing) \right),
  \label{eq:guidance}
  \end{equation}
  where $\gamma$ is the guidance scale and $\varnothing$ represents an empty prompt.  
  At the initial step $t = T$, $z_t$ is sampled from a Gaussian distribution. The latent variable $z_{t-1}$ is then modeled as a Gaussian with parameters derived from $z_t$ and the predicted noise:
  \begin{equation}
  p(z_{t-1} | z_t, p) = \mathcal{N}\left(z_{t-1}; \frac{1}{\sqrt{\alpha_t}} \left( z_t - \frac{\beta_t}{\sqrt{\beta_t}} \hat{\epsilon}_\theta(z_t, t, p) \right), \frac{\bar{\beta}_{t-1} \beta_t^2}{\bar{\beta}_t^2} \mathbf{I} \right),
  \label{eq:posterior}
  \end{equation}
where $\beta_t$ denotes the noise schedule at step $t$, $\alpha_t = 1 - \beta_t$ defines the retained signal ratio, $\bar{\alpha}_t = \prod_{i=1}^{t} \alpha_i$, and $\bar{\beta}_t = \sqrt{1 - \bar{\alpha}_t}$. Following the Markov property, we can compute $\hat{z}_0$ to approximate the denoised latent representation $z_0$~\cite{ho2020denoising}:
  \begin{equation}
  \hat{z}_0 = \frac{1}{\sqrt{\bar{\alpha}_t}} \left( z_t - \bar{\beta}_t \hat{\epsilon}_\theta(z_t, t, p) \right).
  \label{eq:z0}
  \end{equation}
  
 In this way, Eq.(\ref{eq:z0}) mbodies the pivotal inferential step of the diffusion process: recovering a noise-free representation from its corrupted counterpart.  Once that estimate is available, the model channels both the timestep-specific latent code $z_t$ and its refined prediction $\hat{z}_0$ through a latent-to-pixel decoder.  This mapping produces the provisional reconstruction $x_t$, reflecting the system’s current state, and culminates in the fully denoised image $\hat{x}_0$.

  \subsection{Null Space and Range Space: Theoretical Foundation} 

  In our proposed framework, semantic purification is realized through a dual-space projection strategy where the null space is for removing toxic semantics and the range space is for aligning with clean semantics. To ground this approach in mathematical rigor \cite{greub2012linear}, we review the foundational concepts of null space and range space, and explain how they support our token embedding transformations in the context of safe text-to-image generation.
  
  \begin{definition}\label{definition:Null}
  Let \(\boldsymbol{C} \in \mathbb{R}^{d \times k}\) be a matrix formed by stacking embedding vectors associated with a set of known toxic concepts, where \(d\) is the embedding dimension and \(k\) is the number of toxic concept vectors. From the linear algebra such as \cite{greub2012linear}, the \textbf{null space} of \(\boldsymbol{C}\), denoted as \(\text{Null}(\boldsymbol{C})\), is defined as:
  \[
  \text{Null}(\boldsymbol{C}) = \left\{ \boldsymbol{x} \in \mathbb{R}^k \mid \boldsymbol{C}\boldsymbol{x} = \boldsymbol{0} \right\}.
  \]
  This space contains all vectors that, when linearly combined with the columns of \(\boldsymbol{C}\), result in zero. Intuitively, projecting a prompt token embedding into the null space of \(\boldsymbol{C}\) suppresses any components that align with toxic semantics, effectively removing harmful latent information.
  \end{definition}
  
  \begin{definition}\label{definition:Range}
  Let \(\boldsymbol{R} \in \mathbb{R}^{d \times k'}\) be a matrix composed of embedding vectors corresponding to clean or safe concepts. The \textbf{range space} (also known as the \textbf{column space}) of \(\boldsymbol{R}\), denoted as \(\text{Range}(\boldsymbol{R})\), is defined as:
  \[
  \text{Range}(\boldsymbol{R}) = \left\{ \boldsymbol{y} \in \mathbb{R}^d \mid \exists \boldsymbol{x} \in \mathbb{R}^{k'} \text{ such that } \boldsymbol{y} = \boldsymbol{R}\boldsymbol{x} \right\}.
  \]
  This space encompasses all vectors that can be represented as linear combinations of the clean concept embeddings. Projecting a prompt token embedding into the range space of \(\boldsymbol{R}\) reinforces its alignment with safe, desirable semantics.
  \end{definition}
  
  \noindent\textbf{Why Null and Range Spaces Matter for Safe T2I.} These two subspaces play complementary roles in our semantic purification strategy. The null space of \(\boldsymbol{C}\) captures directions that are orthogonal to all toxic concepts; any projection into this space effectively eliminates toxic semantic components. On the other hand, the range space of \(\boldsymbol{R}\) captures the span of clean semantics, and projecting into this space encourages alignment with intended safe meanings.
  
  According to the fundamental theorem of linear algebra, if we consider the full Euclidean space \(\mathbb{R}^d\), the null space and row space of a matrix are orthogonal complements, and likewise for the range space and left null space. Specifically, applying the \textit{Rank-Nullity Theorem} to \(\boldsymbol{C}\), we have:
  \[
  \dim(\text{Null}(\boldsymbol{C})) + \dim(\text{Range}(\boldsymbol{C})) = d,
  \]
  which guarantees that any prompt embedding can be decomposed into components lying in these two orthogonal subspaces.
  
  \noindent\textbf{Connection to Our Method.} In our framework, we leverage this decomposition for semantic filtering and alignment as follows. The projection onto \(\text{Null}(\boldsymbol{C})\) removes the latent toxic semantics by filtering out components in the direction of toxic concept embeddings. The projection onto \(\text{Range}(\boldsymbol{R})\) ensures the prompt remains semantically aligned with safe concepts, enhancing content relevance and generation quality.
  This dual-space view provides a principled geometric perspective for prompt purification. Unlike heuristic-based editing methods, our projection-based approach is grounded in linear algebra and applies directly in the embedding space. It is thus interpretable, efficient, and generalizable across tasks and models, laying a solid theoretical foundation for safe and faithful generation in diffusion models.

  \section{PurifyGen  Framework}
Recent approaches~\citep{gandikota2024unified,gong2024reliable,lu2024mace,chavhan2024conceptprune,yang2024pruning} have demonstrated the effectiveness of weight modification through unlearning or model editing to prevent the generation of harmful (e.g., pornography, self-harm, violence), biased (e.g., racial or social stereotypes, ageism), or otherwise undesirable (e.g., public, copyrighted) visual content in text-to-image generation models. However, these methods have limited flexibility as they require storing individual model weights for each concept to be removed, inherently reduce the backbone model’s generative capabilities through unlearning, and necessitate separate solutions for safe generation across different models (i.e., modified model weights). 

To overcome these challenges, we introduce PurifyGen. Our approach aims to position the prompt close to a list of clean concepts and distant from toxic ones. Initially, we measure the distances between the prompt tokens and these lists to assess the prompt's risk level. We employ the complementary semantic distance to gauge the similarity between the prompt and both the clean and toxic lists. If the prompt's token embeddings are nearer to the clean list than the toxic list, we classify it as safe, allowing direct input into the diffusion model. Otherwise, we transform the prompt embeddings using null and range spaces as semantic regularizers—the former to deter toxic information and the latter to align with safe information. This method enables dual-level alignment, thereby ensuring the generation of safe contexts.

\begin{figure*}[t]
    \centering
    \includegraphics[width=\linewidth]{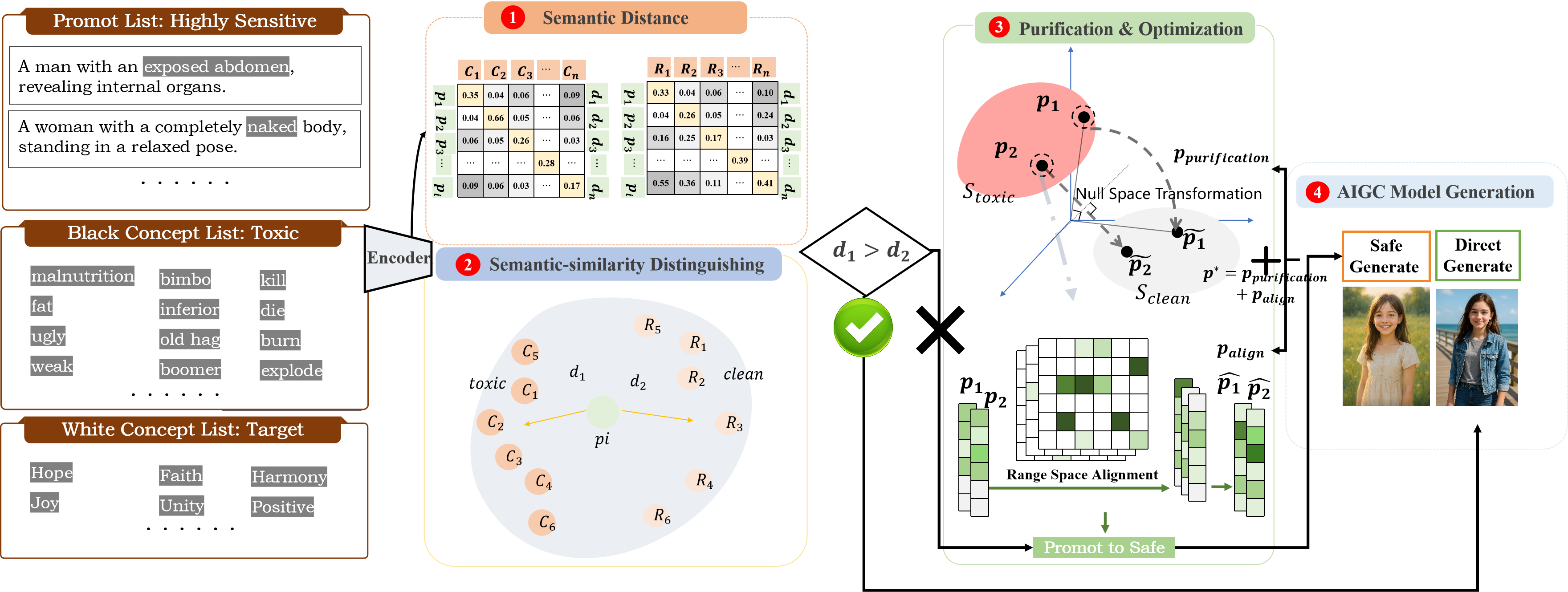}
    \caption{ Illustrations of our framework. Given the white and black checklist and the prompt, we first mine the semantic relationship between prompts with toxic/clean concepts. Later, we determine whether it is a safe, unsafe, or risky prompt, and for the risky ones, we perform the dual-space alignment purification operation. Then, based on our purified prompt embedding, we generate a secure image.}
    \label{fig:concept}
\end{figure*}

\subsection{Discriminate the Safety of Prompts}\label{sec:inference-strategy}
In text-to-image generation tasks, prompts often contain a mix of both toxic and clean content rather than being entirely one or the other. This necessitates the identification of individual tokens within the prompt to determine whether they are safe or pose a risk. For instance, assuming $p_\text{T2I}$=``a man gets killed", only the verb ``killed" is related to ``murder", while ``a'', ``man'', and ``gets'' carry no harmful concept. 

To this end, we propose to detect token embeddings that trigger inappropriate image generation and generate the toxic and clean lists of the concepts as follows:
\begin{equation}
  \begin{aligned}
    \bm{C} &= [\bm{c}_0; \bm{c}_1; \ldots; \bm{c}_{K-1}] \in \mathbb{R}^{D \times K}, \\
    \bm{R} &= [\bm{c}_0'; \bm{c}_1'; \ldots; \bm{c}_{K-1}'] \in \mathbb{R}^{D \times K}, 
  \end{aligned}
\end{equation}
which represents the embedding matrix that denotes the toxic subspace. In this way, each column vector \(\bm{c}_k\in \bm{C}\) corresponds to the embedding of the relevant text associated with the user-defined toxic concept, such as Sexual Acts or Pornography for Nudity concepts. Each column vector \(\bm{c}_k'\in\bm{R}\) corresponds to the embedding of the clean text associated contract with the user-defined toxic concept, such as purity or modesty for non-violence concepts.

Merely checking the cosine similarity between a token and the toxic anchors is unreliable: a high-dimensional embedding can be close to both safe and unsafe directions, or to neither.  
What we truly need is a score that reflects the residual information that survives outside each concept subspace.  In this way,
for the prompt $p$, motivated by previous work \cite{gong2024reliable,safreesafree1}, we identify the complementary semantic distance of $p$ with $\bm{C}$ or $\bm{R}$ as follows:
\begin{equation}
  \mathcal{D}(\bm{C}, \bm{p}) = (\bm{I}-\bm{C})\bm{p}^T,
\end{equation}
\begin{equation}
  \mathcal{D}(\bm{R}, \bm{p}) = (\bm{I}-\bm{R})\bm{p}^T,
\end{equation}
where $\bm{I}$ is the identity matrix. $I-\mathbf{C}$ and $I-\mathbf{R}$ act as lightweight projectors onto the complements of $\mathcal{S}_{\text{tox}}$ and $\mathcal{S}_{\text{safe}}$, respectively.

Intuitively, if the token is closer to the safe concepts than the unsafe concepts in the latent space, the token of the prompt is safe. Otherwise, it is unsafe:
\begin{equation}
  p_i =
  \begin{cases}
  \text{risky}, & \text{if } \mathcal{D}(\bm{C},\bm{p}_i) \leq \mathcal{D}(\bm{R},\bm{p}_i), \\
  \text{safe}, & \text{if } \mathcal{D}(\bm{C},\bm{p}_i) > \mathcal{D}(\bm{R},\bm{p}_i).
  \end{cases}
  \end{equation}
  One can observe that this operation is highly efﬁcient and incurs minimal computational overhead, as $\bm{h}_c$ can be pre-computed and stored for fast inference.  In this way, we classify the prompt as safe or unsafe. Unsafe prompts are filtered, safe prompts are generated directly, and for risky prompts, we purify them as follows.

\subsection{Purification for Prompts}
 
Null space projection is a common technique for removing the influence of a given perturbation. In this work, we adopt the null space projection method to purify the prompt. Specifically, we first compute the null space of the covariance matrix of the prompt, and then project the prompt onto the null space to remove the influence of the perturbation. 
 Following the existing methods for conducting null space projection, we first apply a singular value decomposition (SVD) to \(\bm{C}\):
\begin{equation}
  \{U, \Lambda, (U)^T\} = \text{SVD} \left(\bm{C} \right),
\end{equation}
 where each column in \( U \) is an eigenvector of \( \bm{C} \). Then, we remove the eigenvectors in \( U \) that correspond to non-zero eigenvalues, and define the remaining submatrix as \( \hat{U} \). Based on this, the projection matrix \( P \) can be defined as 
  $\bm{V} = \hat{U}(\hat{U})^T$.
 Then we have the projected prompt as follows:
 \begin{equation}
  \bm{p}^{\text{purification}}_i = (\bm{I} - \bm{V}\bm{V}^T)\bm{p_i},
 \end{equation}

In the above, we propose to leverage the null space projection to avoid toxic information. However, previous work often overlooks the alignment with safe information, and push them into the safe space. 
 Then we design the alignment as follows:

  Let \( \bm{R} \) be a matrix representing safe information, and let \( \bm{P} \) be an input matrix. The projection of \( \bm{p}_i \) onto the space spanned by \( \bm{R} \) ensures alignment with safe information, as defined by:
\begin{equation}
  \bm{p}^{\text{align}}_i = \bm{p}_i \bm{R}^T (\bm{R}^T)^\dagger \bm{R},
\end{equation}
where \( (\bm{R}^T)^\dagger \) is the Moore-Penrose pseudoinverse of \( \bm{R}^T \). This operation ensures that \( \bm{P}^{\text{align}} \) lies in the column space of \( \bm{R} \), thereby aligning \( \bm{P} \) with the safe information represented by \( \bm{R} \).

 Suppose that $\bm{P}$ is the embeddings of the prompt, and $R$ is the mebeddings of the reverse matrix. Then we have the dual-space alignment theory as follows:
\begin{equation}\label{P* for the overall}
  \begin{aligned}
    \bm{P}^*= &\bm{P}_{\text{purification}}+\bm{P}_{\text{align}}\\
   =&\underbrace{(\bm{I} - \bm{V}\bm{V}^T)\bm{P}}_{\text{Avoid Toxic Information}} +  \underbrace{\bm{P} \bm{R}^T(\bm{R}^T)^\dagger \bm{R}}_{\text{Align with Safe Information}}.
  \end{aligned}
\end{equation}
In this way, we generate the new prompt $\bm{P}^*$ from the original prompt $\bm{P}$ by avoiding the toxic information and aligning with the safe information. Different from previous work, we do not need to train a new model to generate the safe prompt. Instead, we can directly generate the safe prompt from the original prompt. 

\subsection{Safe Generation}
We aim to project toxic concept tokens into a safer space to encourage the model to generate appropriate images. In the previous sections, we introduced a dual-space alignment strategy that combines null space projection to remove semantic components associated with toxic concepts and range space alignment to reinforce similarity with clean, safe concepts. This transformation yields a purified prompt embedding $\bm{P}^*$, which is semantically safer and better aligned for responsible generation. However, directly substituting the entire prompt with its purified counterpart may not always be optimal.

In most real-world queries, the objectionable meaning is concentrated in a handful of words; the remainder of the prompt is innocuous yet vital for conveying intent with natural flow. A blanket overhaul, whether by masking every token or applying a uniform replacement, shreds that fluency and expressiveness, particularly when the ``toxic slice" is small.  Equally crude fixes, such as stuffing flagged slots with random vocabulary, zero vectors, or a static ``safe" stub, fracture sentence structure and introduce semantic drift, often yielding gibberish generations.

To avoid these pitfalls, we adopt an adaptive, token-wise purification routine. Let $\mathbf{P}_i$ be the embedding of the $i$-th token. Using the Semantic-based risk test from Section~\ref{sec:inference-strategy}, we label each token as \emph{risky} or \emph{benign}.  Risky embeddings are swapped for sanitised counterparts $\mathbf{P}_i^{*}$ generated by our dual-space aligner, while benign ones remain untouched to preserve fidelity.  The resulting prompt embedding is therefore
\begin{equation}
  \mathbf{P}_i^{\text{safe}} = 
\begin{cases}
\mathbf{P}_i^{*}, & \text{if the token is risky},\\[4pt]
\mathbf{P}_i,        & \text{otherwise}.
\end{cases}
\end{equation}

Compared to prior methods that indiscriminately suppress or remove toxic cues, our approach offers a more fine-grained and context-aware solution.This design ensures minimal perturbation to the input prompt, offering a balance between semantic safety and content preservation.

\begin{table*}[t]
  \caption{The ASR and generation quality comparison with several popular safe T2I generation methods. Risk-D Ability denortes the risk-discrimination ability. MMA denotes MMA-Diffusion. The ones marked with * indicate training-based methods, which are not included in the performance comparison for fairness. The best results are \textbf{bolded}.}
   \label{tab:t2i_main}
   \small
   \centering
   \resizebox{0.99\textwidth}{!}{
   \setlength{\tabcolsep}{0.3mm}
   \begin{tabu}{lccc|ccccc|ccc}
       \toprule
       & \multicolumn{3}{c|}{\textbf{Method Properties}} & \multicolumn{5}{c|}{\textbf{Attack Success Rate}} & \multicolumn{3}{c}{\textbf{COCO Metrics}}\\
       \cmidrule(lr){2-4} \cmidrule(lr){5-9} \cmidrule(lr){10-12}
       \textbf{Method}& 
       \begin{tabular}[c]{@{}c@{}}No Weights\\Modification\end{tabular} & 
       \begin{tabular}[c]{@{}c@{}}Training\\-Free\end{tabular} & 
       \begin{tabular}[c]{@{}c@{}}Risk-D\\Ability\end{tabular} &
       I2P $\downarrow$ & P4D $\downarrow$ & Ring-A-Bell $\downarrow$ & MMA $\downarrow$ & UDA $\downarrow$ &
       FID $\downarrow$ & CLIP $\uparrow$ & TIFA $\uparrow$\\
       \midrule
       SD-v1.4 &-&-&-& 0.178 & 0.987 & 0.831 & 0.957 & 0.697 & 163.5 & 31.3 & 0.3166 \\
       \midrule
       {ESD~\citep{gandikota2023erasing}}$^*$ &\xmark&\xmark&\xmark& 0.140 & 0.750 & 0.528 & 0.873 & 0.761 & - & 29.7 & - \\
       SA~\citep{heng2023selective}$^*$&\xmark&\xmark&\xmark& 0.062 & 0.623 & 0.329 & 0.205 & 0.268 & 210.2 & 29.6 & 0.290 \\
       CA~\citep{kumari2023ablating}$^*$&\xmark& \xmark&\xmark& 0.178 & 0.927 & 0.773 & 0.855 & 0.866 & 158.1 & 30.02 & 0.293 \\
       {MACE~\citep{lu2024mace}}$^*$&\xmark&\xmark&\xmark& 0.023 & 0.146 & 0.076 & 0.183 & 0.176 & 195.6 & 29.4 & 0.265 \\
       {SDID~\citep{li2024self}}$^*$&\xmark&\xmark&\xmark& 0.270 & 0.933 & 0.696 & 0.907 & 0.697 & 89.4 & 29.5 & 0.299 \\
       \midrule
       UCE~\citep{gandikota2024unified}&\xmark&\textcolor{Green}{\cmark}&\xmark& 0.103 & 0.667 & 0.331 & 0.867 & 0.430 & 120.7 & 30.33 & 0.300 \\
       RECE~\citep{gong2024reliable}&\xmark&\textcolor{Green}{\cmark}&\xmark& 0.064 & 0.381 & 0.134 & 0.675 & 0.655 & 145.2 & 29.9 & 0.294 \\
       \midrule
       SLD-Medium~\citep{schramowski2023safe} &\textcolor{Green}{\cmark}&\textcolor{Green}{\cmark}&\xmark& 0.142 & 0.934 & 0.646 & 0.942 & 0.648 & \textbf{117.3} & 30.01 & 0.292 \\
       SLD-Strong~\citep{schramowski2023safe}&\textcolor{Green}{\cmark}&\textcolor{Green}{\cmark}&\xmark& 0.131 & 0.861 & 0.620 & 0.920 & 0.570 & 152.6 & 29.4 & 0.286 \\
       SLD-Max~\citep{schramowski2023safe} &\textcolor{Green}{\cmark}&\textcolor{Green}{\cmark}&\xmark& 0.115 & 0.742 & 0.570 & 0.837 & 0.479 & 188.1 & 28.6 & 0.267 \\
       SAFREE \cite{safreesafree1} &\textcolor{Green}{\cmark}&\textcolor{Green}{\cmark}&\xmark& 0.034 & 0.412 & 0.295 & 0.585 & 0.282 & 137.4 & 29.05 & 0.291 \\
       \midrule
       Ours &\textcolor{Green}{\cmark}&\textcolor{Green}{\cmark}&\textcolor{Green}{\cmark}&\textbf{0.028} & \textbf{0.377} & \textbf{0.126} & \textbf{0.553} & \textbf{0.174} & 134.2 & \textbf{30.47} & \textbf{0.301} \\
       \bottomrule
   \end{tabu}
   }
\end{table*}
\section{Experimental Results}
\subsection{Experimental Setup}
Our experimental backbone is \textit{Stable Diffusion} v1.4 (SD-v1.4)~\citep{rombach2022high}, the same engine that underpins several recent safety-oriented studies~\citep{gandikota2023erasing,gandikota2024unified,gong2024reliable}.  
To probe each method’s resilience, we barrage it with adversarial prompts sourced from five red-teaming pipelines: I2P~\citep{schramowski2023safe}, P4D~\citep{chin2024prompting4debugging}, Ring-a-Bell~\citep{tsai2024ring}, MMA-Diffusion~\citep{yang2024mma}, and UnlearnDiff~\citep{zhang2023generate}.  
Following the evaluation recipe in~\citep{gandikota2023erasing}, we additionally assess the capacity to strip copyrighted or trademarked artistic styles.  
All these styles can be reproducible by SD-v1.4, making them a stringent test bed for style-suppression techniques.

\begin{table}
    \caption{{Comparison of Artist Concept Removal tasks}: Famous (left)  and Modern artists (right).}\label{tab:t2i_art}\vspace{-0.1in}
    \centering
    \setlength{\tabcolsep}{0.5mm}
    \resizebox{0.48\textwidth}{!}{
    \begin{tabu}{l|cccc|cccc}
    \toprule
    &\multicolumn{4}{c}{\textbf{Remove ``Van Gogh"}}&\multicolumn{4}{c}{\textbf{Remove ``Kelly McKernan"}}\\ 
    \cmidrule(lr){2-5}
    \cmidrule(lr){6-9}
    \textbf{Method}&
    \textbf{LPIPS}$_e \uparrow$ &\textbf{LPIPS}$_u\downarrow$ & \textbf{Acc}$_e \downarrow$& \textbf{Acc}$_u\uparrow$&
    \textbf{LPIPS}$_e\uparrow$ &\textbf{LPIPS}$_u\downarrow$ & \textbf{Acc}$_e\downarrow$& Acc$_u\uparrow$\\
    \midrule
    SD-v1.4&-&-&0.95&0.95&-&-&0.80&0.83\\
    \midrule
    CA&0.30&0.13&0.65&0.90&0.22&0.17&0.50&0.76\\
    RECE&0.31&0.08&0.80&0.93&0.29&0.04&0.55&0.76\\
    UCE&0.25&{0.05}&0.95&0.98&0.25&{0.03}&0.80&0.81\\
    \midrule
    SLD-Medium&0.21&0.10&0.95&0.91&0.22&0.18&0.50&0.79\\
    SAFREE& {0.42}&{0.31}& {0.35}&0.85& {0.40}&{0.39}& {0.40}&0.78\\
    Ours&\textbf{0.48}&{0.25}&\textbf{0.31}&0.87&\textbf{0.44}&\textbf{0.33}&\textbf{0.37}&0.79\\
    \bottomrule
    \end{tabu}
    }
    \end{table}

    \begin{table}[t]
        \centering
        \caption{Comparison of methods on sensitive content metrics. $\downarrow$ indicates lower is better.}
        \footnotesize
        \setlength{\tabcolsep}{0.5mm}
        \resizebox{0.48\textwidth}{!}{
        \begin{tabu}{lccccc}
        \toprule
        Methods & Violence $\downarrow$ & {Terrorism $\downarrow$} & {Racism $\downarrow$} & {Sexual $\downarrow$} & {Animal Abuse $\downarrow$} \\
        \midrule
        {ZeroScopeT2V} & 71.68 & 76.00 & 73.33 & 51.51 & 66.66 \\
        {ZeroScopeT2V + Ours} & \textbf{48.61} & \textbf{50.09} & \textbf{53.47} & \textbf{13.26} & \textbf{35.14} \\
        \midrule
        {CogVideoX-5B} & 80.12 & 76.00 & 73.33 & 75.75 & 92.59 \\
        {CogVideoX-5B + Ours} & \textbf{49.23} & \textbf{54.00} & \textbf{63.24} & \textbf{28.40} & \textbf{44.27} \\
        \bottomrule
        \end{tabu}
        }
        \label{text-to-video generation}
        \end{table}

\noindent\textbf{Baselines.} Our method is compared against several training-free approaches: SLD~\citep{schramowski2023safe} and UCE~\citep{gandikota2024unified}. For training-based methods, including ESD~\citep{gandikota2023erasing}, SA~\citep{heng2023selective}, CA~\citep{kumari2023ablating}, MACE~\citep{lu2024mace}, SDID~\citep{li2024self}, SAFREE \cite{safreesafree1}, and RECE~\citep{gong2024reliable}. 

\noindent\textbf{Evaluation Metrics.} We quantify how effectively each defence blocks illicit nudity requests by reporting the \textit{attack success rate} (ASR) introduced in~\citep{gong2024reliable}; lower is safer.  For benign generations we sample 1\,000 COCO-30k captions~\citep{lin2014microsoft} and compute three complementary metrics: Fréchet Inception Distance (FID)~\citep{heusel2017gans} for distributional realism, the CLIP similarity score for semantic alignment, and TIFA~\citep{hu2023tifa} for text–image faithfulness.  When evaluating artist-signature removal, visual discrepancy between the output and an unaltered reference is gauged with LPIPS~\citep{zhang2018unreasonable}.  In addition, we recast the task as a multi-choice quiz: an autonomous GPT-4o agent~\citep{gpt4o} is asked to guess the artist underlying each generated picture.  A higher error rate implies more successful style suppression.

\subsection{Experimental Results}
\textbf{Overall Performance.}   
We subjected each safety mechanism to a rigorous suite of red-team prompts and tallied the fraction of queries that still slipped through, reported as the attack success rate (ASR).  As summarised in Table~\ref{tab:t2i_main}, \textit{PurifyGen} posts the lowest ASR across every attack variant, eclipsing all other training-free baselines.  Compared with the strongest prior defences, such as I2P, MMA-Diffusion, and UnlearnDiff. Our method drives the success rate sharply downward, underscoring a step-change in robustness against adversarial provocations.

\begin{table}[t]
    \centering
    \caption{Adversarial Prompt and CoCo evaluation results. MMA denotes MMA-Diffusion. }
    \label{Adversarial Prompt and CoCo Evaluation Results}
    \resizebox{0.48\textwidth}{!}{
    \begin{tabu}{cccccccc}
    \toprule
    \textbf{DSP} & \textbf{PP} & \textbf{PA} & \multicolumn{3}{c}{\textbf{Adversarial Prompt}} & \multicolumn{2}{c}{\textbf{CoCo}} \\
    \cmidrule(lr){4-6} \cmidrule(lr){7-8}
    & & & \textbf{P4D} $\downarrow$ & \textbf{MMA} $\downarrow$ & \textbf{Ring-a-Bell} $\downarrow$ & \textbf{FID} $\downarrow$ & \textbf{CLIP} $\uparrow$ \\
    \midrule
    $\times$ & \checkmark & \checkmark & 0.425& 0.613& 0.285& 139.2& 28.0\\
    \checkmark & \checkmark & $\times$ & 0.412& 0.595& 0.295& 137.4& 29.1\\
    \checkmark& $\times$&\checkmark & 0.428& 0.583& 0.295& 136.9& 29.6\\
    \checkmark& \checkmark & \checkmark & \textbf{0.377}& \textbf{0.55}& \textbf{0.126}& \textbf{134.2}& \textbf{30.1}\\
    \midrule
    \end{tabu}}
    \end{table}

\begin{table}[t]
\centering
\label{tab:combined}
\centering
\caption{Model Efficiency Comparison. All experiments are tested on a single A6000, 100 steps, and with a setting that removes the nudity concept. }\label{tab:t2i_efficiency}\vspace{-0.05in}
\setlength{\tabcolsep}{1.mm}
\resizebox{0.48\textwidth}{!}{
\begin{tabular}{lccc}
\toprule
\textbf{Method} & \multicolumn{1}{c}{\begin{tabular}[c]{@{}c@{}}\textbf{Training/Editing}\\ \textbf{Time (s)}\end{tabular}} & \multicolumn{1}{c}{\begin{tabular}[c]{@{}c@{}}\textbf{Inference}\\ \textbf{Time (s/sample)}\end{tabular}}  & \multicolumn{1}{c}{\begin{tabular}[c]{@{}c@{}}\textbf{Model}\\ \textbf{Modification (\%)}\end{tabular}} \\ \midrule
ESD &$\sim$4500& 6.78 & 94.65\\
CA    &$\sim$484&5.94& 2.23 \\
UCE & $\sim$1 &6.78 & 2.23 \\ 
RECE  &$\sim$3& 6.80 & 2.23\\ \midrule
SLD-Max & 0 & 9.82 & 0 \\
SAFREE & 0 & 9.85& 0\\
Ours & 0 & 9.83& 0\\
\bottomrule
\end{tabular}
}
\end{table}

\begin{figure*}[t]
  \centering
  \begin{subfigure}{0.45\linewidth}
      \includegraphics[width=\linewidth]{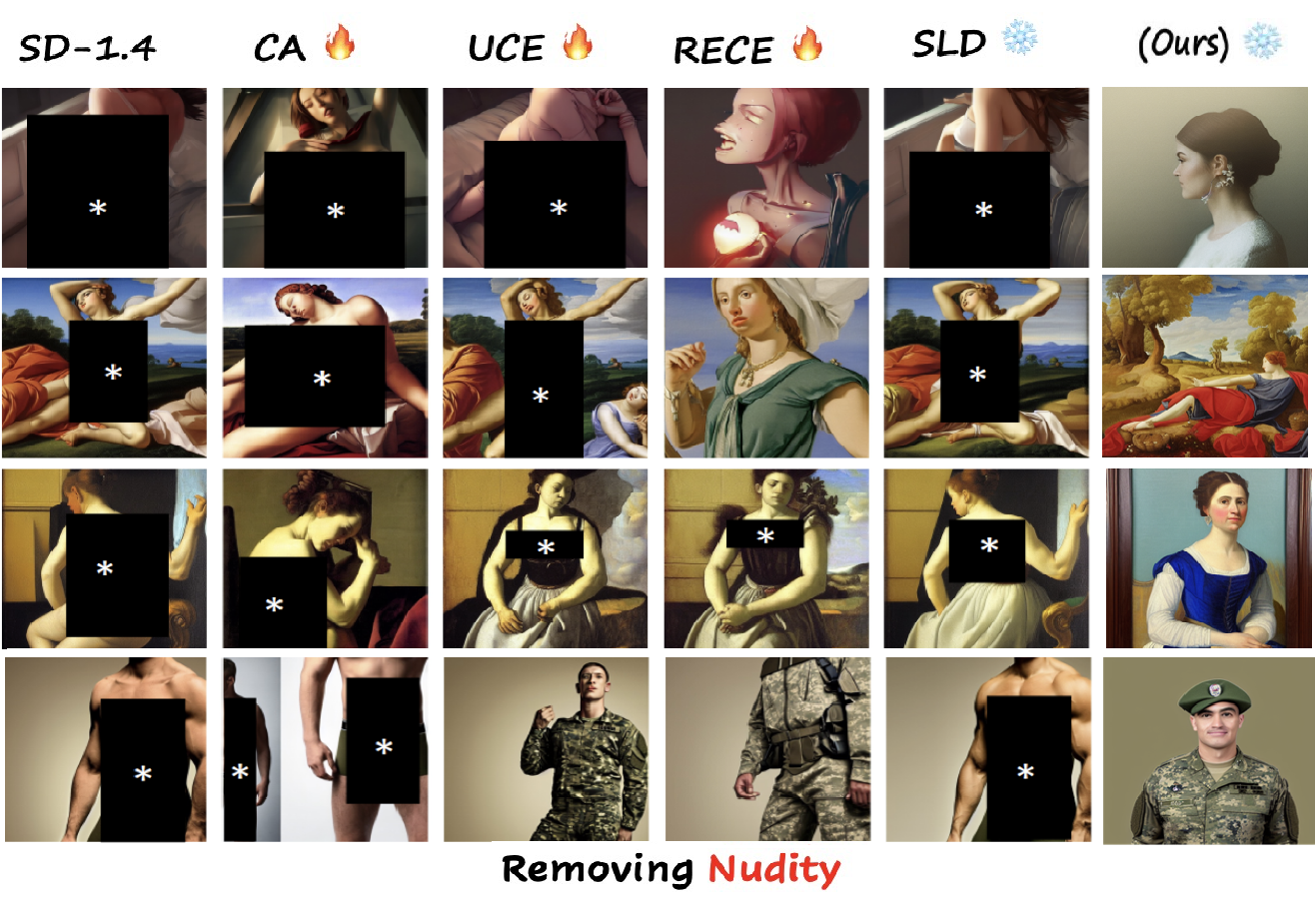}
      \caption{ Case 1}
  \end{subfigure}
  \hfill
  \begin{subfigure}{0.45\linewidth}
      \includegraphics[width=\linewidth]{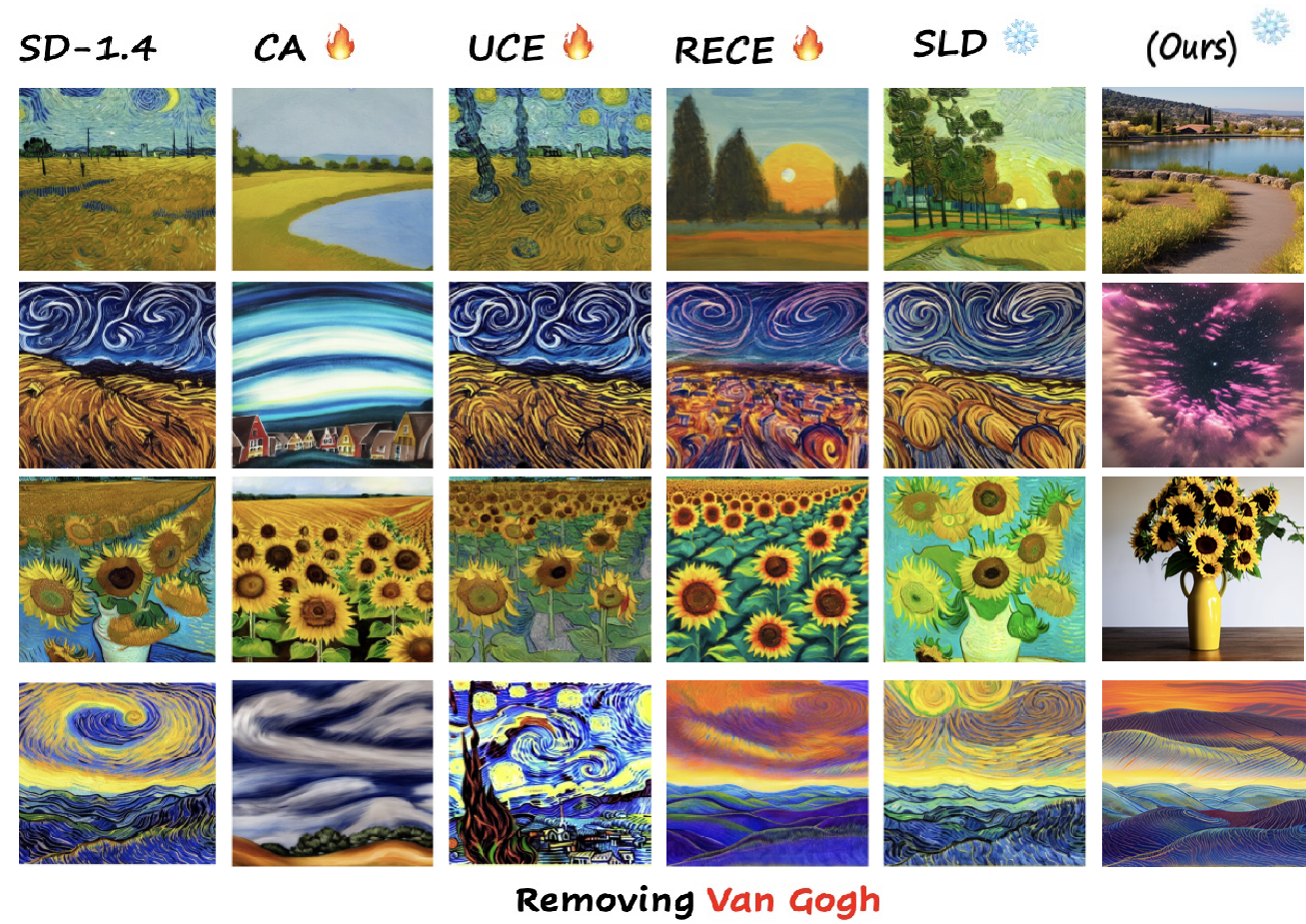}
      \caption{Case 2}
  \end{subfigure}
  \caption{Generated examples of PurifyGen and safe T2I baselines. The two cases demonstrate nudity and  Van Gogh concept removal.}
  \label{fig:vis-nudity}
\end{figure*}

\noindent\textbf{Comparison to Training-based Methods.} In our comparison with training-based approaches, we find that PurifyGen achieves performance on par with these techniques. As demonstrated in Table \ref{tab:t2i_main} and Figure  \ref{fig:vis-nudity}, while methods such as SA and MACE demonstrate strong safety capabilities, they often cause significant degradation in image quality due to the extensive modifications to the SD weights. These alterations typically introduce noticeable distortions, making these methods impractical for real-world use. In contrast, as demonstrated on the COCO-30k dataset, PurifyGen provides similar safeguarding performance while maintaining high-quality image generation without training.

 Figure~\ref{fig:concept} illustrates qualitative comparisons across multiple baselines and our proposed method PurifyGen for two representative concept removal scenarios: \textit{nudity} (Case 1) and \textit{Van Gogh style} (Case 2). These examples provide compelling evidence of PurifyGen's ability to precisely remove undesired content while maintaining semantic fidelity and image quality.
In Case 1, where the goal is to eliminate nudity, we observe that baselines such as CA, UCE, and RECE either fail to completely suppress explicit content or introduce significant visual artifacts and distortions. For example, CA and UCE often preserve recognizable body parts while crudely masking others, leading to incoherent or ambiguous images. On the other hand, SLD attempts to suppress nudity but occasionally over-censors the image, introducing structural inconsistencies or damaging the prompt's intent. In contrast, PurifyGen successfully removes the nudity while preserving the context, pose, and artistic style of the original image. The output images remain faithful to the prompt's broader semantics, such as clothing style or character composition, highlighting the precision of our token-wise purification mechanism.

In Case 2, we task each method with removing stylistic attributes associated with the "Van Gogh" concept. Again, baseline methods struggle to decouple style from content. For instance, CA, UCE, and RECE only partially reduce Van Gogh's characteristic brushstroke patterns, often leaving residual stylization or transforming the prompt into low-quality outputs. SLD, while better at abstraction, sometimes introduces unnatural textures or degrades the vibrancy of the image. By contrast, PurifyGen consistently generates outputs that are stylistically distinct from Van Gogh's artwork while maintaining the original scene layout and compositional structure. For example, our method transforms Van Gogh-style sunflowers into realistic floral arrangements, preserving subject matter without carrying over unwanted stylistic influence.

\noindent\textbf{Ablation Studies.} We assess the contributions of three key components of PurifyGen in Table \ref{Adversarial Prompt and CoCo Evaluation Results} using SD-v1.4. We denote the component including discriminating the risky of prompt as DSP. We denote the purification component as PP. We denote the alignment component as PA. In this way, one can observe that all variants substantially reduce the performance from adversarial prompts, confirming the effectiveness of our components. 

With all three modules active the system achieves the uniformly best numbers: 0.377 on P4D, 0.550 on MMA, and a mere 0.126 on Ring-a-Bell, up to a 55 \% reduction in the hardest setting, while simultaneously improving image realism (FID 134.2) and text-image correspondence (CLIP 30.1).  The monotonic improvement from single- to dual-component variants, and the decisive leap with the full trio, demonstrates that DSP pinpoints the threat, PP neutralises it, and PA restores coherence; the trio works synergistically to shrink the adversarial surface and polish visual quality, showing that safety enforcement need not compromise generative fidelity in a training-free pipeline.

\subsection{Further Experimantal Analysis}
\noindent\textbf{Evaluating PurifyGen on Artist Concept Removal Tasks.}
In Table \ref{tab:t2i_art}, we show that PurifyGen outperforms baseline methods in terms of LPIPS$_e$ and LPIPS$_u$ scores. We focus on comparing with the training-free methods. The improved performance is attributed to our approach's denoising process, which is guided by a coherent, projected conditional embedding within the input space. To evaluate whether the generated art styles are accurately removed or preserved, we frame these tasks as a multiple-choice question-answering problem, moving beyond simple feature-level distance measures. In this context, Acc$_e$ and Acc$_u$ represent the average accuracy of removed and retained artist styles predicted by GPT-4o based on the corresponding text prompts. As shown in Table \ref{tab:t2i_art}, PurifyGen demonstrates strong performance in removing targeted artist concepts, while baseline methods struggle to eliminate key representations of the target artists.

\begin{figure}[t]
  \centering
    \centering
    \includegraphics[width=\linewidth]{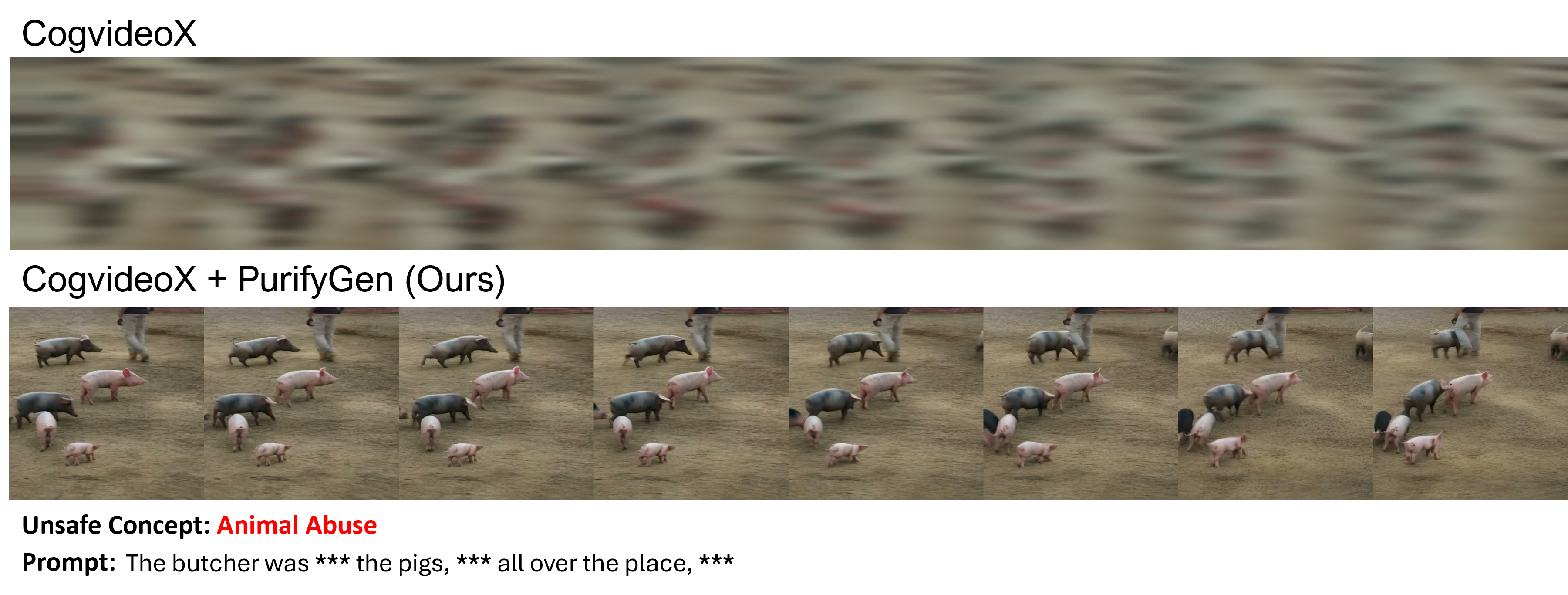}
  \caption{Experimental results with CogVideoX.}
  \label{fig:text-video}
\end{figure}

\noindent\textbf{Transfer to Text--to--Video.}  
We next test whether PurifyGen's safeguards extend beyond still imagery by grafting it onto two leading T2V diffusion systems: ZeroScopeT2V, whose denoiser is UNet-based, and CogVideoX-5B, built around a Diffusion Transformer.  Both hybrids are evaluated on the SafeSora benchmark~\cite{dai2024safesora}, which spans a comprehensive roster of unsafe themes.  Results (Table~\ref{text-to-video generation}, Fig.~\ref{fig:text-video}) reveal a steep contraction of harmful frames across all five risk classes.  Concretely, ZeroScopeT2V's sexual-content rate falls from 51.51 \% to 13.26 \%, with violent scenes dropping from 71.68 \% to 48.61 \%.  An analogous pattern appears for CogVideoX‐5B.  These outcomes confirm that PurifyGen generalises smoothly to moving-image synthesis and remains effective across disparate generative backbones, positioning it as a scalable safety layer for both video and image pipelines.

\noindent\textbf{Efficiency of PurifyGen.}
In this section, we compare the efficiency of various methods, including the training-based ESD/CA, which relies on online optimization and loss to update model weights, and the training-free UCE/RECE, which modify model attention weights through closed-form edits. Our method (PurifyGen), similar to SLD, is training-free and based on filtering without requiring any alterations to the diffusion model weights. As shown in Table \ref{tab:t2i_efficiency}, while UCE/RECE allows for rapid model editing, they still require additional time for model updates. In contrast, PurifyGen does not necessitate any model edits or modifications, offering greater flexibility for model development across varying conditions while maintaining competitive generation speeds. Based on the results presented in Table \ref{tab:t2i_efficiency}, PurifyGen demonstrates superior performance in concept safeguarding, generation quality, and flexibility, making it a highly efficient solution.

\section{Conclusion}
In this paper, we propose PurifyGen, an effective solution for safe text-to-image generation that requires neither model retraining nor additional data. By leveraging dual-space semantic transformations, it ensures both the removal of harmful content and the preservation of prompt intent. Its token-wise purification strategy minimizes collateral degradation of safe information. Experimental results demonstrate its strong generalization and superior safety performance across diverse benchmarks. Moreover, PurifyGen is compatible with a wide range of diffusion backbones, making it a practical and scalable choice for real-world deployment. Future work will explore extending this purification framework to multi-modal generation tasks and interactive prompt refinement.

\section{Acknowledgments}
This work was supported by the Science and
Technology Innovation 2030-Key Project under Grant 2021ZD0201404.

\clearpage

\bibliographystyle{ACM-Reference-Format}
\balance
\bibliography{main}
\clearpage

\end{document}